\documentclass[10pt,twocolumn,letterpaper]{article}

\usepackage[pagenumbers]{cvpr} % To force page numbers, e.g. for an arXiv version

\usepackage{graphicx}
\usepackage{amsmath}
\usepackage{amssymb}
\usepackage{booktabs}
\usepackage{multirow}
\usepackage{bbding}
\usepackage[table]{xcolor}
\usepackage{enumitem}
\usepackage{amsfonts,bm}
\usepackage[normalem]{ulem}
\usepackage{subcaption}

\usepackage[pagebackref,breaklinks,colorlinks]{hyperref}
\DeclareMathAlphabet{\mathsfit}{\encodingdefault}{\sfdefault}{m}{sl}
\SetMathAlphabet{\mathsfit}{bold}{\encodingdefault}{\sfdefault}{bx}{n}
\newcommand{\tens}[1]{\bm{\mathsfit{#1}}}

\usepackage[capitalize]{cleveref}
\crefname{section}{Sec.}{Secs.}
\Crefname{section}{Section}{Sections}
\crefname{table}{Tab.}{Tabs.}
\Crefname{table}{Table}{Tables}

\def\confName{CVPR}
\def\confYear{2022}

\begin{document}

\title{Cascade Transformers for End-to-End Person Search}

\author{Rui Yu$^{1,2}$\thanks{Rui Yu's work on this paper was done when he was a summer intern at Kitware.},\hspace{0.75cm}Dawei Du$^{1}$,\hspace{0.75cm}Rodney LaLonde$^{1}$,\hspace{0.75cm}Daniel Davila$^{1}$,\\Christopher Funk$^{1}$,\hspace{0.75cm}Anthony Hoogs$^{1}$,\hspace{0.75cm}Brian Clipp$^{1}$\\
$^1$Kitware, Inc., NY \& NC, USA,\hspace{0.75cm}$^2$Pennsylvania State University, PA, USA\\
{\tt\small \url{https://github.com/Kitware/COAT}}
}
\maketitle

\newcommand{\comment}[1]{
\noindent\textcolor{red}{\itshape{\textbf{#1}} }
}

%%%%%%%%% ABSTRACT
\begin{abstract}
The goal of person search is to localize a target person from a gallery set of scene images, which is extremely challenging due to large scale variations, pose/viewpoint changes, and occlusions. In this paper, we propose the Cascade Occluded Attention Transformer (COAT) for end-to-end person search. Our three-stage cascade design focuses on detecting people in the first stage, while later stages simultaneously and progressively refine the representation for person detection and re-identification. At each stage the occluded attention transformer applies tighter intersection over union thresholds, forcing the network to learn coarse-to-fine pose/scale invariant features. Meanwhile, we calculate each detection's occluded attention to differentiate a person's tokens from other people or the background. In this way, we simulate the effect of other objects occluding a person of interest at the token-level. Through comprehensive experiments, we demonstrate the benefits of our method by achieving state-of-the-art performance on two benchmark datasets.
\end{abstract}

%%%%%%%%% BODY TEXT
\section{Introduction}
\label{sec:intro}
Person search aims to localize a particular target person from a gallery set of scene images, which is an extremely difficult fine-grained recognition and retrieval problem. A person search system must both \textit{generalize} to separate people from the background, and \textit{specialize} to discriminate identities from each other. 
\begin{figure}[t]
    \centering
    \includegraphics[width=\linewidth]{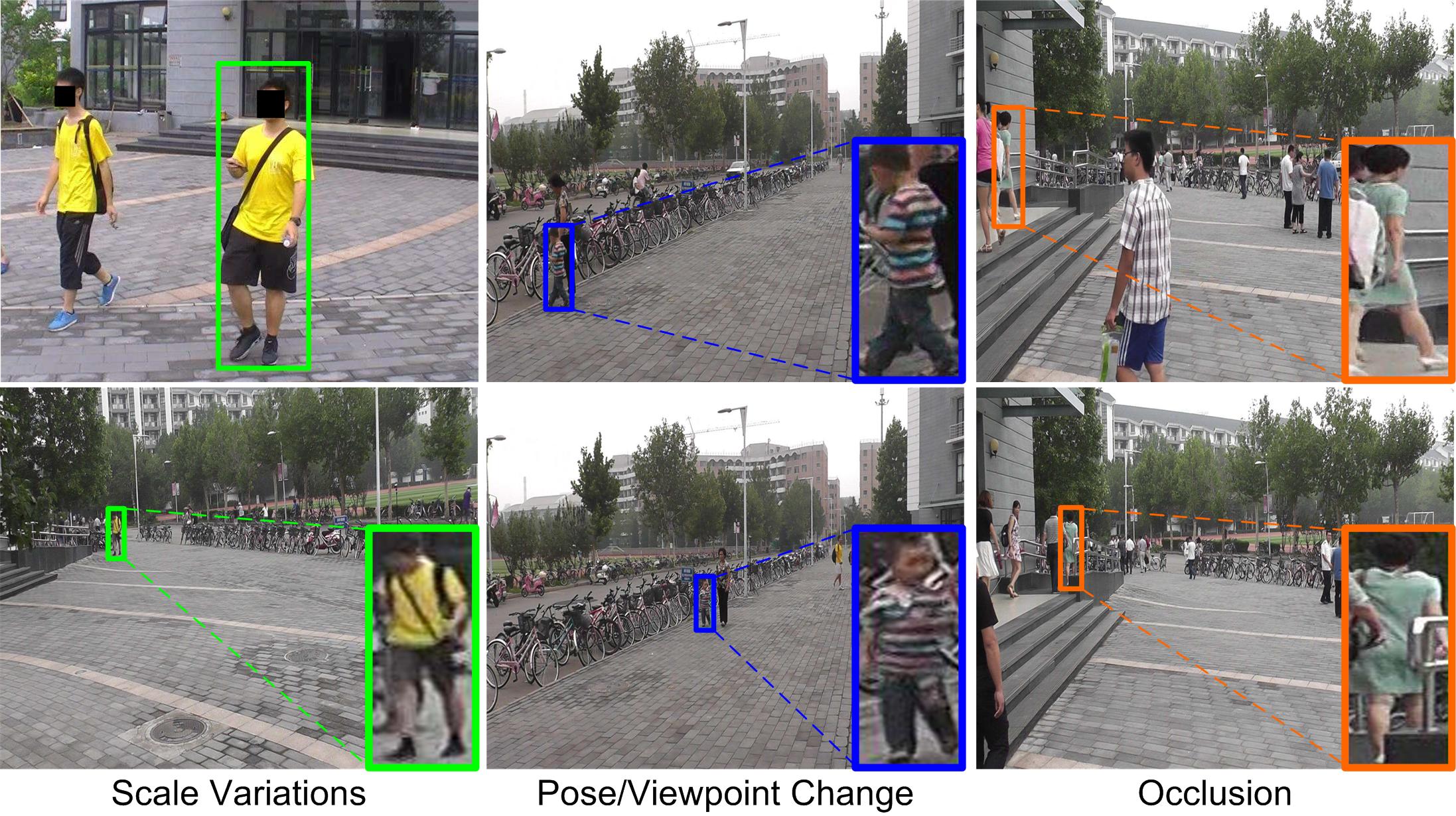}
    \caption{Main challenges of person search, \eg, scale variations, pose/viewpoint change, and occlusion. The boxes with the same color represent the same ID. For better viewing, we highlight the small-scale individuals at bottom-right corners. \label{fig:intro}}
    \vspace{-20pt}
\end{figure}

In real-world applications, person search systems must detect people across a wide variety of image sizes and re-identify people despite large changes in resolution and viewpoint.
To this end, modern person search methods, either two-step or one-step (\ie, end-to-end), consist of reliable person detection and discriminative feature embedding learning.
Two-step methods~\cite{DBLP:conf/cvpr/ZhengZSCYT17,DBLP:conf/eccv/ChenZOYT18,DBLP:conf/eccv/LanZG18,DBLP:conf/iccv/HanYZTZGS19,DBLP:conf/cvpr/DongZST20,DBLP:conf/cvpr/WangMCSC20} conduct person re-identification (ReID) on cropped person patches found by a separate object detector.
In contrast, end-to-end methods~\cite{DBLP:conf/cvpr/XiaoLWLW17,DBLP:conf/eccv/ChangHSLYH18,DBLP:conf/cvpr/YanZNZXY19,DBLP:conf/cvpr/ZhongWZ20,DBLP:journals/corr/abs-2103-11617,DBLP:conf/aaai/LiM21} jointly solve the detection and ReID sub-problems in a more efficient, multi-task learning framework.
However, as shown in Figure~\ref{fig:intro}, they still suffer from three main challenges:
\begin{itemize}[leftmargin=*,nosep]
    \item \textit{There is a conflict in feature learning between person detection and ReID.} Person detection aims to learn features which generalize across people to distinguish people from the background, while ReID aims to learn features which do \textit{not} generalize across people but distinguish people from each other. Previous works follow a ``ReID first''~\cite{DBLP:journals/corr/abs-2103-11617} or ``detection first''~\cite{DBLP:conf/aaai/LiM21} principle to give priority to one subtask over the other. However, it is difficult to balance the importance of two subtasks in different situations when relying on either strategy.
    
    \item \textit{Significant scale or pose variations increase identity recognition difficulty}; see Figure~\ref{fig:intro}. Feature pyramids or deformable convolutions~\cite{DBLP:conf/eccv/LanZG18, DBLP:conf/aaai/HanZGSY21,DBLP:journals/corr/abs-2103-11617} have been used to solve scale, pose or viewpoint misalignment in feature learning. However, simple feature fusion strategies may introduce additional background noise in feature embeddings, resulting in inferior ReID performance.
    
    \item \textit{Occlusions with background objects or other people make appearance representations more ambiguous}, as shown in Figure~\ref{fig:intro}. The majority of previous person search methods focus on holistic appearance modeling of people by anchor-based~\cite{DBLP:conf/aaai/LiM21} or anchor-free~\cite{DBLP:journals/corr/abs-2103-11617} methods. Despite the improvement of person search accuracy, these are prone to fail with complex occlusions.
\end{itemize}

\begin{figure}[t]
    \centering
    \includegraphics[width=1.0\linewidth]{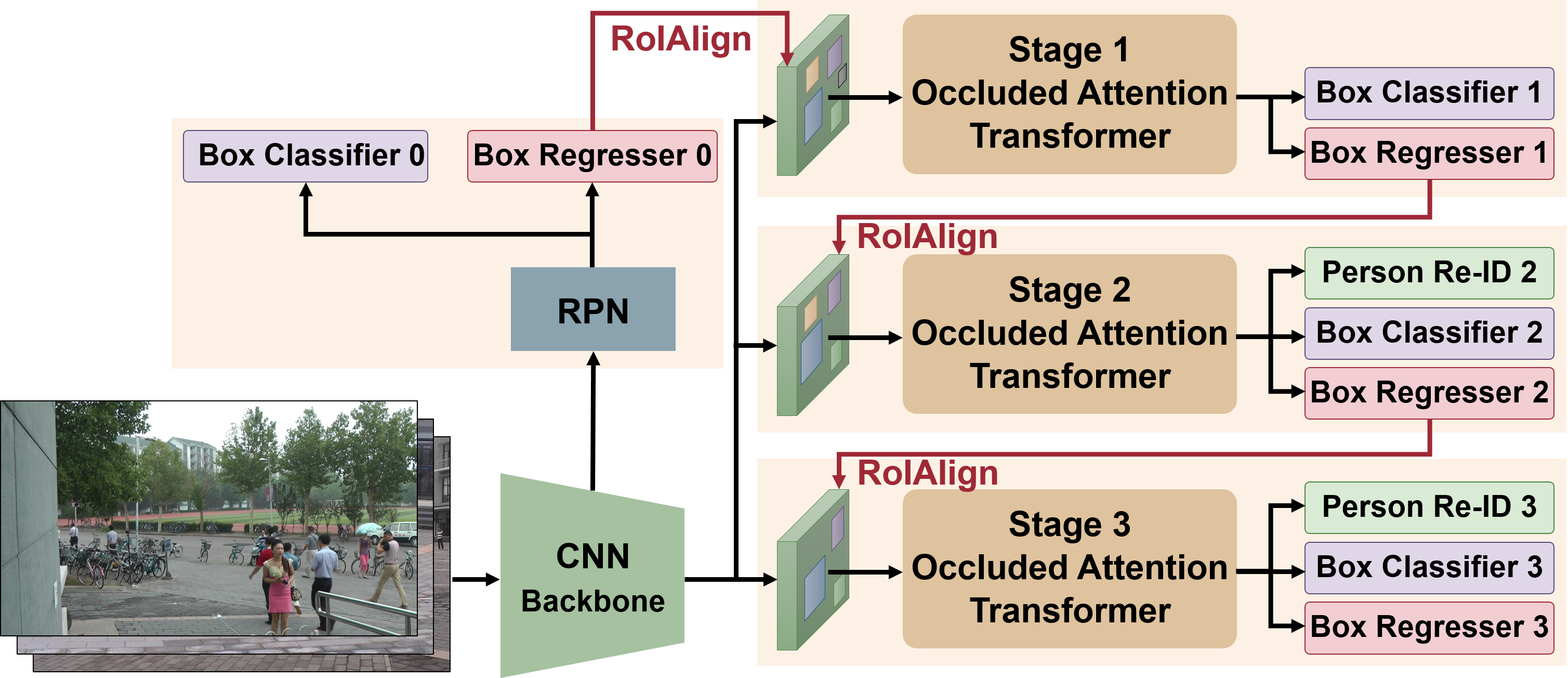}
    \caption{Our proposed cascade framework for person search. \label{fig:framework}
    \vspace{-20pt}}
    
\end{figure}

To deal with the aforementioned challenges, as shown in Figure~\ref{fig:framework}, we propose a new Cascade Occluded Attention Transformer (COAT) for end-to-end person search.
First, inspired by Cascade R-CNN~\cite{DBLP:conf/cvpr/CaiV18}, we refine the person detection and ReID quality by a coarse-to-fine strategy in three stages. The first stage focuses on discriminating people from background (detection), but crucially, is not trained to discriminate people from each other (ReID) with a ReID loss.  Later stages include both detection and ReID losses. This design improves detection performance (see Section~\ref{sec:ablation}), as the first stage can generalize across people without having to discriminate between persons. Subsequent stages simultaneously refine the previous stages' bounding box estimates and identity embeddings (see Table~\ref{tab:ablationCascade}). 
Second, we apply multi-scale convolutional transformers at each stage of the cascade. The base feature maps are split into multiple slices corresponding to different scales. The transformer attention encourages the network to learn embeddings on the discriminative parts of each person for each scale, helping overcome the problem of region misalignment.
Third, we augment the transformer's learned feature embeddings with an occluded attention mechanism that synthetically mimics occlusions . We randomly mix-up partial tokens of instances in a mini-batch, and learn the cross-attention among the token bank for each instance.  This trains the transformer differentiate tokens from other foreground and background detection proposals.
Experiments on the challenging CUHK-SYSU~\cite{DBLP:conf/cvpr/XiaoLWLW17} and PRW~\cite{DBLP:conf/cvpr/ZhengZSCYT17} datasets show that the proposed network outperforms state-of-the-art end-to-end methods, especially in terms of the cross-camera setting on the PRW dataset.

{\noindent {\bf Contributions.}} 1) To our knowledge, we propose the first cascaded transformer-based framework for end-to-end person search. The progressive design effectively balances person detection and ReID and the transformers help attend to scale and pose/viewpoint changes. 2) We improve performance with an occluded attention mechanism in the multi-scale transformer that generates discriminative fine-grained person representations in occluded scenes. 3) Extensive experiments on two datasets show the superiority of our method over existing person search approaches.

\section{Related Work}
{\noindent {\bf Person Search.}} 
Person search methods can be roughly grouped into two-step and end-to-end approaches. Two-step methods~\cite{DBLP:conf/eccv/ChenZOYT18,DBLP:conf/eccv/LanZG18,DBLP:conf/iccv/HanYZTZGS19,DBLP:conf/cvpr/DongZST20,DBLP:conf/cvpr/WangMCSC20} combine a person detector (\eg, Faster R-CNN~\cite{DBLP:journals/pami/RenHG017}, RetinaNet~\cite{DBLP:conf/iccv/LinGGHD17}, or FCOS~\cite{DBLP:conf/iccv/TianSCH19}) and a person ReID model sequentially. 
For example, Wang~\etal~\cite{DBLP:conf/cvpr/WangMCSC20} build a person search system including an identity-guided query detector followed by a detection results adapted ReID model. 
On the other hand, end-to-end methods~\cite{DBLP:conf/cvpr/XiaoLWLW17,DBLP:conf/cvpr/ChenZYS20,DBLP:journals/corr/abs-2103-11617,DBLP:conf/aaai/LiM21} integrate the two models into a unified framework for better efficiency.
Chen~\etal~\cite{DBLP:conf/cvpr/ChenZYS20} share detection and ReID features but decompose them in the polar coordinate system in terms of radial norm and angle.
Yan~\etal~\cite{DBLP:journals/corr/abs-2103-11617} propose the first anchor-free person search method, which tackles the misalignment issues in different levels (\ie, scale, region, and task).
Recently, Li and Miao~\cite{DBLP:conf/aaai/LiM21} share the stem representations of person detection and ReID, but solve the two subtasks by two-head networks sequentially.
In contrast, inspired by Cascade R-CNN~\cite{DBLP:conf/cvpr/CaiV18}, our method follows an end-to-end strategy that balances person detection and ReID progressively via a three-stage cascade framework.

{\noindent {\bf Visual Transformers in Person ReID.}}
Based on the original transformer model~\cite{DBLP:conf/nips/VaswaniSPUJGKP17} for natural language processing, Vision Transformer (ViT)~\cite{DBLP:conf/iclr/DosovitskiyB0WZ21} is the first pure transformer network to extract features for image recognition. 
CNNs are widely adopted to extract base features and so reduce the scale of training data required for a pure transformer approach. Luo~\etal~\cite{DBLP:journals/tmm/LuoJFZ20} develop a spatial transformer network to sample an affined image from the holistic image to match a partial image.
Li~\etal~\cite{DBLP:journals/corr/abs-2106-04095} propose the part-aware transformer to perform occluded person Re-ID through diverse part discovery.
Zhang~\etal~\cite{zhang2021hat} introduce a transformer-based
feature calibration to integrate large scale features as a global prior.
Our paper is the first in the literature to perform person search with multi-scale convolutional transformers . It not only learns discriminative ReID features but also distinguishes people from the background in a cascade pipeline.

{\noindent {\bf Attention Mechanism in Transformers.}}
Attention mechanism plays a crucial role in transformers. Recently, many ViT variants ~\cite{DBLP:journals/corr/abs-2103-14899,DBLP:journals/corr/abs-2106-05786,DBLP:journals/corr/abs-2101-11986,DBLP:journals/corr/abs-2102-04378} have computed discriminative features using a variety of token attention methods.
Chen~\etal~\cite{DBLP:journals/corr/abs-2103-14899} propose a dual-branch transformer with a cross-attention based token fusion module to combine two scales of patch features.
Lin~\etal~\cite{DBLP:journals/corr/abs-2106-05786} alternate attention in the feature map patches for local representation and attention on the single channel feature map for global representation.
Yuan~\etal~\cite{DBLP:journals/corr/abs-2101-11986} introduce the tokens-to-token process to gradually tokenize images to tokens while preserving structural information. 
He~\etal~\cite{DBLP:journals/corr/abs-2102-04378} rearrange the transformer layers' patch embeddings via shift and patch shuffle operations. 
Unlike these methods that rearrange features within an instance, the proposed occluded attention module considers token cross-attention between either positive or negative instances from the mini-batch. Thus our method learns to differentiate tokens from other objects by synthetically mimicking occlusions.  

\section{Cascade Transformers}
\begin{figure*}[t]
\centering
\includegraphics[width=0.95\linewidth]{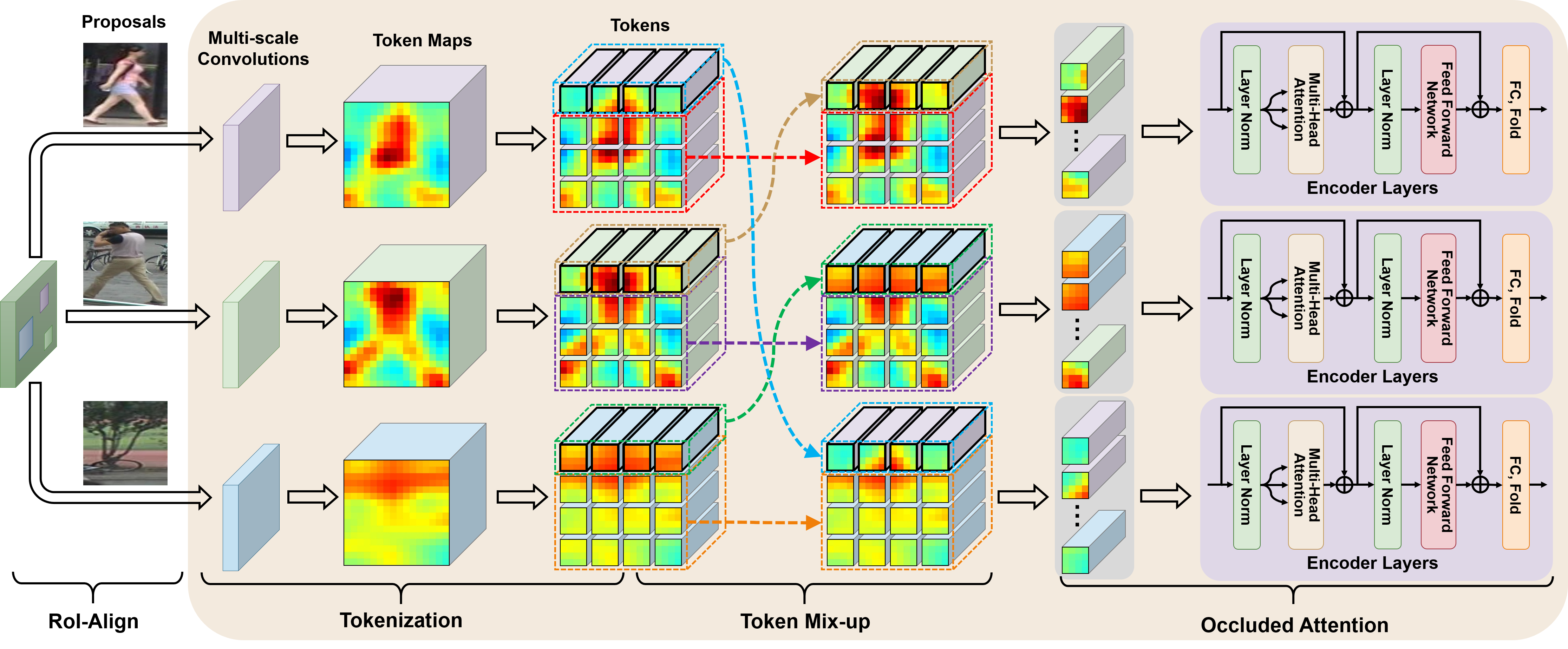}
\caption{Architecture of occluded attention transformer. The randomly selected regions for token exchange are the same within one mini-batch. For clarity, we only show three instances in a mini-batch and occluded attention for one scale. Best view in color. \label{fig:transformer}}
\vspace{-4mm}
\end{figure*}

As discussed in previous works~\cite{DBLP:journals/corr/abs-2103-11617, DBLP:conf/aaai/HanZGSY21, DBLP:conf/aaai/LiM21},  person detection and person ReID have conflicting goals. Hence, it is difficult to jointly learn discriminative unified representations for the two subtasks on the top of the backbone network. 
Similar to Cascade R-CNN~\cite{DBLP:conf/cvpr/CaiV18}, we decompose feature learning into sequential steps in $T$ stages of multi-scale transformers. That is, each head in the transformer refines the detection and ReID accuracy of the predicted objects stage-by-stage. Thus we can progressively learn coarse-to-fine unified embeddings.

Nevertheless, in the case of occlusions by other people, objects or the background, the network may suffer from noisy representations of the target identity. To this end, we develop the occluded attention mechanism in the multi-scale transformer to learn an occlusion-robust representation.
As shown in Figure~\ref{fig:framework}, our network is based on the Faster R-CNN object detector backbone with Region Proposal Network (RPN). However, we extend the framework by introducing a cascade of occluded attention transformers (see Figure~\ref{fig:transformer}), trained in an end-to-end manner.

\subsection{Coarse-to-fine Embeddings}
% base feature maps
After extracting the $1024$-dim stem feature maps from the ResNet-50~\cite{DBLP:conf/cvpr/HeZRS16} backbone, we use the RPN to generate region proposals. For each proposal, the RoI-Align operation~\cite{DBLP:journals/pami/RenHG017} is applied to pool an $h\times w$ region as the base feature maps $\mathcal{F}$, where $h$ and $w$ denote the height and width of the feature maps respectively, and $c$ is the number of channels.

% cascade feature
Afterwards, we employ a multi-stage cascade structure to learn embeddings for person detection and ReID. The output proposals of the RPN are used at the first stage for re-sampling both positive and negative instances. The box outputs of the first stage are then adopted as the inputs of the second stage, and so forth. At each stage $t$, the pooled feature map of each proposal is sent to the convolutional transformers for that stage.
To obtain high-quality instances, the cascade structure imposes progressively more strict stage-wise constraints. In practice, we increase the intersection-over-union (IoU) thresholds $u_t$ gradually. The transformers at each stage are followed by three heads, like NAE~\cite{DBLP:conf/cvpr/ChenZYS20}, including a person/background classifier, a box regressor, and a ReID discriminator. Note that we remove the ReID discriminator at the first stage to focus the network on first detecting all people in the scene before refinement.

\subsection{Occluded Attention Transformer} \label{sec:occTransformer}
In the following, we describe the details of the occluded attention transformers, shown in Figure~\ref{fig:transformer}. 

{\noindent {\bf Tokenization.}}
Given the base feature map $\mathcal{F}\in\mathbb{R}^{h\times w \times c}$, we tokenize it for transformer input at different scales. For multi-scale representation, we first split $\mathcal{F}$ channel-wise into $n$ slices, $\bar{\mathcal{F}}\in\mathbb{R}^{h\times w \times \hat{c}}$, where $\hat{c}=\frac{c}{n}$ to deal with each scale of token.
In contrast to ViT~\cite{DBLP:conf/iclr/DosovitskiyB0WZ21} with its tokenization of large image patches, our transformer leverages a series of convolutional layers to generate tokens based on the sliced feature maps $\bar{\mathcal{F}}$. Our method benefits from CNNs' inductive biases
and learns the CNN's local spatial context. The different scales are realized by different sizes of convolutional kernels.

After converting the sliced feature maps $\bar{\mathcal{F}}\in\mathbb{R}^{h\times w \times \hat{c}}$ to the new token map $\hat{\mathcal{F}}\in\mathbb{R}^{\hat{h}\times\hat{w}\times \hat{c}}$ by one convolutional layer, we flatten it into tokens inputs $\mathbf{x}\in\mathbb{R}^{\hat{h}\hat{w}\times \hat{c}}$ for one instance. The number of tokens calculated as
\begin{equation}
N=\frac{\hat{h}\hat{w}}{d^2}=\frac{\lfloor\frac{h+2p-k}{s}+1\rfloor \times \lfloor\frac{w+2p-k}{s}+1\rfloor}{d^2},
\label{equ:tokens}
\end{equation}
where we have the kernel size $k$, stride $s$, and padding $p$ for the convolutional layer. $d$ is the patch size of each token.

{\noindent {\bf Occluded attention.}}
To handle occlusions, we introduce a new token-level occluded attention mechanism into the transformers to mimic occlusions found in real applications.
Specifically, we first collect the tokens from all the detection proposals in a mini-batch, denoted as \textit{token bank} $\mathbf{X}=\{\mathbf{x}_1, \mathbf{x}_2, \cdots, \mathbf{x}_P\}$, where $P$ is the number of detection proposals in the batch at each stage. Since the proposals the from RPN contain positive and negative examples, the token bank is composed of both foreground person parts and background objects.
We exchange tokens among the token bank, based on the same exchange index set $\mathcal{M}$ for all the instances. As shown in Figure \ref{fig:transformer}, the exchanged tokens correspond to a semantically consistent but randomly selected sub-regions in the token maps. Each exchanged token is denoted as
\begin{equation}
\mathbf{x}_i=\{\mathbf{x}_i(\bar{\mathcal{M}}), \mathbf{x}_j(\mathcal{M})\}, \quad i=1,2,\cdots,P, i\neq j,
\label{equ:exchange}
\end{equation} 
where $\mathbf{x}_j$ denotes another sample randomly selected from the token bank. $\bar{\mathcal{M}}$ indicates the complementary set of $\mathcal{M}$, \ie, $\mathbf{x}_i = \mathbf{x}_i(\bar{\mathcal{M}})\bigcup \mathbf{x}_i(\mathcal{M})$.
% multi-scale attention
Given the exchanged token bank $\mathbf{X}$, we compute the multi-scale self-attention among them, as shown in Figure \ref{fig:transformer}.
In terms of each scale of tokens, we run two sub-layers of the transformers (\ie, Multi-head Self-Attention (MSA) and a Feed Forward Network (FFN) as in~\cite{DBLP:conf/nips/VaswaniSPUJGKP17}).
Specifically, the mixed tokens $\mathbf{x}$ are transformed into \textit{query} matrices $\tens{Q}\in\mathbb{R}^{\hat{h}\hat{w}\times\hat{c}}$, \textit{key} matrices $\tens{K}\in\mathbb{R}^{\hat{h}\hat{w}\times\hat{c}}$, and \textit{value} matrices $\tens{V}\in\mathbb{R}^{\hat{h}\hat{w}\times\hat{c}}$ by three individual fully connected (FC) layers.
We can further compute multi-head attention and the weighted sum over all values as
\begin{equation}
\mathrm{MSA}(\tens{Q},\tens{K},\tens{V}) = \mathrm{softmax}(\frac{\tens{Q}\tens{K}^\mathrm{T}}{\sqrt{\hat{c}/m}})\tens{V},
\label{equ:attention}
\end{equation}
where we split queries, keys, and values into $m$ heads for more diversity, \ie, from tensor with the size of $\hat{h}\hat{w}\times\hat{c}$ to $m$ pieces with the size of $\hat{h}\hat{w}\times\frac{\hat{c}}{m}$. The independent attention outputs are then concatenated and linearly transformed into the expected dimension.
Following the MSA module, the FFN module nonlinearly transforms each token to enhance its representation ability.
The enhanced feature is then projected to the size of $\hat{h}\times \hat{w}\times \hat{c}$ as the transformer's output.

Finally, we concatenate the outputs of the $n$ scales of transformers to original spatial size $\hat{h}\times \hat{w}\times c$. Note that there is a residual connection outside each transformer.
After the global average pooling (GAP) layer, the extracted features are fed into subsequent heads for box regression, person/background classification, and person re-identification.

{\noindent {\bf Relations to concurrent works.}} \label{sec:discuss}
There are two concurrent ViT based works~\cite{DBLP:journals/corr/abs-2103-14899, DBLP:journals/corr/abs-2102-04378} in different fields. Chen~\etal~\cite{DBLP:journals/corr/abs-2103-14899} develop a multi-scale transformer including two separate branches with small-patch and large-patch tokens. The two-scale representation is learned based on a cross-attention token fusion module, where a single token for each branch is treated as a query to exchange information with other branches. 
Instead, we leverage a series of convolutional layers with different kernels to generate multi-scale tokens. Finally, we concatenate the enhanced feature maps corresponding to each scale in specific slice of the transformers. 

To deal with occlusion and misalignment in person ReID, He~\etal~\cite{DBLP:journals/corr/abs-2102-04378} shuffle person part patch embeddings and re-group them, each group of which contains several random patch embeddings of an individual instance.
In contrast, our method first exchanges partial tokens of instances in a mini-batch, and then calculate the occluded attention based on mixed tokens. Thus the final embeddings partially cover the target person with extracted features from a different person or a background object, yielding more occlusion-robust representations.

\subsection{Training and Inference}
In the training phase, the proposed network is trained end-to-end for person detection and person ReID. The person detection loss $\mathcal{L}_\text{det}$ consists of regression and classification loss terms. The former is a Smooth-L1 loss of regression vectors between ground-truth and foreground boxes, while the latter computes the cross-entropy loss of predicted classification probabilities of the estimated boxes. 

To supervise person ReID, we use the classic non-parametric Online Instance Matching (OIM) loss~\cite{DBLP:conf/cvpr/XiaoLWLW17} $\mathcal{L}_\text{OIM}$, which maintains a lookup table (LUT) and a circular queue (CQ) to store the features of all the labeled and unlabeled identities from recent mini-batches, respectively. We can efficiently compute the cosine similarities between the samples in the mini-batch and LUT/CQ for embedding learning. Moreover, inspired by~\cite{DBLP:conf/cvpr/0004GLL019}, we add another cross-entropy loss function $\mathcal{L}_\text{ID}$ to predict the identities of people for an additional ID-wise supervision. In summary, we train the proposed COAT by using the following multi-stage loss:
\begin{equation}
\mathcal{L}=\sum_{t=1}^T{\mathcal{L}_\text{det}^t+\mathbb{I}(t>1)(\lambda_\text{OIM}\mathcal{L}_\text{OIM}^t+\lambda_\text{ID}\mathcal{L}_\text{ID}^t)},
\label{equ:all_loss}
\end{equation}
where $t\in\{1,2,\dots,T\}$ denotes the index of the stage and $T$ is the number of cascade stages. The coefficients $\lambda_\text{OIM}$ and $\lambda_\text{ID}$ are used to balance the OIM and ID loss terms. $\mathbb{I}(t>1)$ is the indicator function to indicate that we do not consider person ReID loss at the first stage.

In the inference phase, we replace the occluded attention mechanism with the classic self-attention module in the transformers by removing the token mix-up step in Figure~\ref{fig:transformer}. We output the detection bounding boxes with corresponding embeddings at the last stage and use NMS operations to remove redundant boxes.

\section{Experiments}
All experiments are conducted in PyTorch with one NVIDIA A100 GPU. For a fair comparison with prior works, we use the first four residual blocks (\texttt{conv1}$\sim$\texttt{conv4}) of ResNet-50~\cite{DBLP:conf/cvpr/HeZRS16} as the backbone and resize the images to $900\times1500$ as the input.

\subsection{Datasets}
We evaluate our method on two publicly available datasets.
The \textbf{CUHK-SYSU} dataset~\cite{DBLP:conf/cvpr/XiaoLWLW17} annotates $8,432$ identities and $96,143$ bounding boxes in $18,184$ images. The default gallery size is set as $100$ for the $2,900$ testing identities in $6,978$ images.
The \textbf{PRW} dataset~\cite{DBLP:conf/cvpr/ZhengZSCYT17} collects data from $6$ cameras, including $932$ identities and $43,110$ pedestrian boxes in $11,816$ frames. PRW is divided into a training set with $5,704$ frames and $482$ identities and a testing set with $2,057$ query persons in $6,112$ frames.

We follow the standard evaluation metrics for person search~\cite{DBLP:conf/cvpr/ZhengZSCYT17,DBLP:conf/cvpr/XiaoLWLW17}. A box is matched if the overlap ratio between the predicted and ground-truth boxes with the same identity is more than $0.5$ IoU. 
For person detection, we use Recall and Average Precision (AP). 
For person ReID, we use the mean Average Precision (mAP) and cumulative matching characteristics (top-1) scores. 

\subsection{Implementation Details}
Similar to Cascade R-CNN~\cite{DBLP:conf/cvpr/CaiV18}, we use $T=3$ stages in the cascade framework, where $128$ detection proposals are extracted per image for each stage. Following~\cite{DBLP:conf/cvpr/XiaoLWLW17,DBLP:conf/cvpr/ChenZYS20,DBLP:conf/aaai/LiM21}, the scale of the base feature map is set as $h=w=14$. 
The index of exchanging tokens in Eq.~\eqref{equ:exchange} is set as the random horizontal or vertical strip in the token map. The number of heads in Eq.~\eqref{equ:attention} is set as $m=8$.
The IoU thresholds $u_t$ for detection are set as $0.5, 0.6, 0.7$ for the three sequential stages. The kernel sizes of the convolutional layers to compute the tokens are set as $k=\{1\times1,3\times3\}$ for the three stages, with corresponding strides $s=\{1,1\}$ and paddings $p=\{0,1\}$ to guarantee the same size of output feature maps. Due to the small feature size, we set $d=1$ in Eq.~\eqref{equ:exchange}, \ie, conducting pixel-wise tokenization.
The CQ size of the OIM loss is set as $5,000$ and $500$ for CUHK-SYSU and PRW respectively. 
The loss weights in Eq.~\eqref{equ:all_loss} are set as $\lambda_\text{OIM}=\lambda_\text{ID}=0.5$. 

We use the SGD optimizer with momentum $0.9$ to train our model for $15$ epochs, with an initial learning rate warming up to $0.003$ during the first epoch, being reduced by a factor of $10$ at the $10$-th epoch. At the inference phase, we use NMS with $0.4/0.4/0.5$ threshold to remove redundant boxes detected by the first/second/third stage. 
%-------------------------------------------------------------------------
\subsection{Ablation Studies}
\label{sec:ablation}
We conduct a series of ablation studies on the PRW dataset~\cite{DBLP:conf/cvpr/ZhengZSCYT17} to analyze our design decisions.

\begin{table}[t]
\begin{center}
\small
\vspace{-4mm}
\setlength{\tabcolsep}{10pt}
%\resizebox{\columnwidth}{!}{
\begin{tabular}{ccc|cc}
\hline
\textbf{Stage1} & \textbf{Stage2} & \textbf{Stage3} & \textbf{mAP}  &\textbf{top-1}\\
\hline
\multicolumn{5}{c}{(a) \textit{w/o Transformers}:}\\
\hline
\XSolidBrush &  &  &43.5  &81.2 \\
\dag\XSolidBrush & \XSolidBrush &  &47.7  &84.6 \\
\dag\XSolidBrush & \dag\XSolidBrush & \XSolidBrush &48.4  &85.2 \\
\dag\XSolidBrush & \XSolidBrush & \XSolidBrush &49.5  &85.5 \\
\XSolidBrush & \XSolidBrush & \XSolidBrush &47.2  &84.9 \\
\hline
\multicolumn{5}{c}{(b) \textit{w/ Transformers}:}\\
\hline
\Checkmark & & &43.3  &78.7 \\
\dag\Checkmark &\Checkmark & &50.8  &84.9 \\
\dag\Checkmark &\dag\Checkmark &\Checkmark &51.3  &85.5 \\
\rowcolor{gray!25}
\dag\Checkmark &\Checkmark &\Checkmark &\textbf{53.3}  &\textbf{87.4} \\
\Checkmark &\Checkmark &\Checkmark &50.3  &84.0 \\
\hline
\multicolumn{5}{c}{(c) \textit{IoU Thresholds}:}\\
\hline
$0.5$ & $0.5$ & $0.5$ &52.5  &86.0 \\
$0.6$ & $0.6$ & $0.6$ &52.6  &86.2 \\
$0.7$ & $0.7$ & $0.7$ &51.0  &85.5 \\
$0.5$ & $0.6$ & $0.6$ &52.6  &86.3 \\
\rowcolor{gray!25}
$0.5$ & $0.6$ & $0.7$ &\textbf{53.3}  &\textbf{87.4} \\
\hline
\end{tabular}
\caption{Comparison with different cascade variants of COAT on PRW~\cite{DBLP:conf/cvpr/ZhengZSCYT17}. ``\XSolidBrush'' means using the same ResNet block (\texttt{conv5}) as~\cite{DBLP:conf/cvpr/XiaoLWLW17,DBLP:conf/cvpr/ChenZYS20,DBLP:conf/aaai/LiM21}, while ``\Checkmark'' means using the proposed transformers at each stage. ``\dag'' means the heads without the ReID loss. Gray highlighting indicates the parameters selected for our final system.
\label{tab:ablationCascade}\vspace{-30pt}}
%}
\end{center}
\end{table}

{\noindent {\bf Contribution of cascade structure.}}
To show the cascade structure's contribution, we evaluate coarse-to-fine constraints in terms of the number of cascade stages and IoU thresholds.

First, we replace the occluded attention transformer with the same ResNet block (\texttt{conv5}) as~\cite{DBLP:conf/cvpr/XiaoLWLW17,DBLP:conf/cvpr/ChenZYS20,DBLP:conf/aaai/LiM21} at each stage. 
As shown in Table~\hyperref[tab:ablationCascade]{\ref*{tab:ablationCascade}(a)}, the cascade structure significantly improves person search accuracy when adding more stages, \ie, from $43.5\%$ to $49.5\%$ in mAP and $81.2\%$ to $85.5\%$ in top-1 accuracy. As we introduce the proposed occluded attention transformer, the performance is further improved (see Table~\hyperref[tab:ablationCascade]{\ref*{tab:ablationCascade}(b)}), which demonstrates our occluded attention transformer's effectiveness .

Moreover, the increasing IoU thresholds $u_t$ in the cascade design improve person search performance. As reported in Table~\hyperref[tab:ablationCascade]{\ref*{tab:ablationCascade}(c)}, equal IoU thresholds at each stage produce lower accuracy than our method. For example, more false positives or false negatives are introduced if $u_t=0.5$ or $u_t=0.7$. In contrast, our method can select detection proposals with increasing quality for better performance, \ie, generating more candidate detections in the first stage and only highly-overlapping detections by the third stage.

\begin{figure}[t]
\centering
\includegraphics[width=\linewidth]{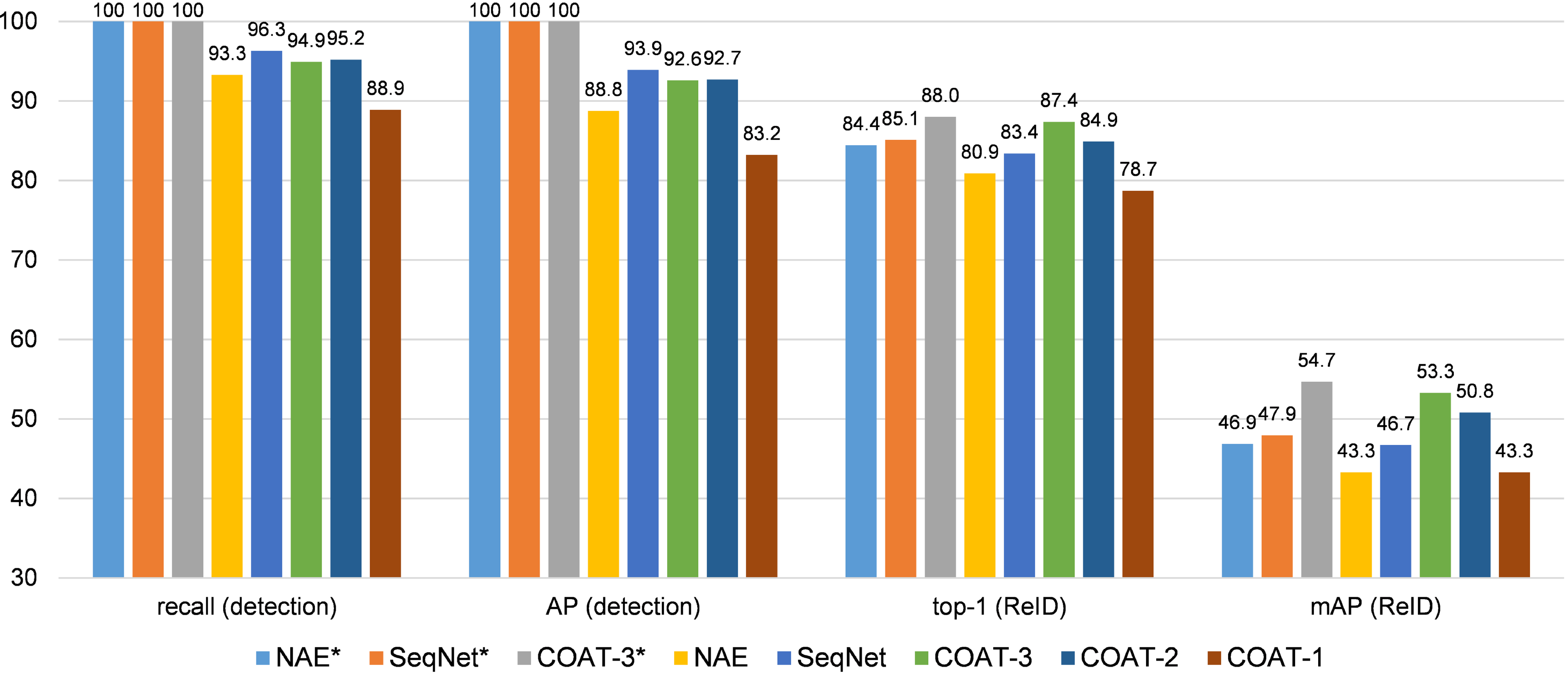}
 \caption{Detection and person search results for COAT and two compared methods on PRW, both with (person ReID only) and without (person search) ground-truth detection boxes being provided. $\ast$ denotes the oracle results using the ground-truth boxes. \label{fig:relation}}
\vspace{-8mm}
\end{figure}

{\noindent {\bf Relations between person detection and ReID.}}
% relation between person detection and ReID
As discussed in the introduction, there is a conflict between person detection and ReID. In Figure~\ref{fig:relation}, we explore the relationship between the two subtasks. We compare our COAT with state-of-the-art NAE~\cite{DBLP:conf/cvpr/ChenZYS20} and SeqNet~\cite{DBLP:conf/aaai/LiM21}, which share the same Faster R-CNN detector. We also construct three COAT variants with different stages, \ie, COAT-$t$, where $t=1,2,3$ denotes the number of stages.
When looking solely at person ReID rather than person search, \ie, when ground-truth detection boxes are given, COAT outperforms the two competitors with an over $3\%$ gain in top-1 and over $6\%$ gain in mAP. 
Meanwhile, our is slightly worse in person detection accuracy than SeqNet~\cite{DBLP:conf/aaai/LiM21}.
These results indicate that our improved ReID performance comes from coarse-to-fine person embeddings rather than more precise detections.

We also observe that the person detection performance is improved from $t=1$ to $t=2$ but then slightly reduced with $t=3$. We speculate that this is because, when trading-off person detection and ReID, our method focuses more on learning discriminative embeddings for person ReID, while slightly sacrificing detection performance.

% influence of person detection and ReID in each stage
In addition, from Table~\hyperref[tab:ablationCascade]{\ref*{tab:ablationCascade}(a)(b)}, note that the COAT variant with ReID loss in the first stage performs worse than our method ($50.3$ vs. $53.3$ for mAP). Simultaneously learning a discriminative representation for person detection and ReID is extremely difficult. Therefore, we remove the ReID disciminator head at Stage $1$ in the COAT method (\textit{c.f.} Figure~\ref{fig:framework}).
If we continue removing the ReID discriminator at the second stage, the ReID performance is reduced by $\sim2\%$ in mAP.
This shows the ReID embeddings do benefit from multi-stage refinement.

{\noindent {\bf Comparison with other attention mechanisms.}}
To verify the effectiveness of our occluded attention mechanism in the transformer, we apply the recently proposed Jigsaw~\cite{DBLP:journals/corr/abs-2102-04378} and CrossViT~\cite{DBLP:journals/corr/abs-2103-14899} in our method. 
As discussed in Section~\ref{sec:discuss}, Jigsaw Patch~\cite{DBLP:journals/corr/abs-2102-04378} is used to generate robust ReID features by shift and patch shuffle operations. CrossViT~\cite{DBLP:journals/corr/abs-2103-14899} is a dual-branch transformer to learn multi-scale features. It is also noteworthy that they leverage large image patches as the input for pure vision transformers. We also evaluate the COAT variant a vanilla self-attention mechanism, denoted as vanilla attention. 

In Table~\ref{tab:cross}, CrossViT~\cite{DBLP:journals/corr/abs-2103-14899} focuses on exchanging information between two scales of tokens, achieving inferior mAP. The results show that Jigsaw~\cite{DBLP:journals/corr/abs-2102-04378} also hurts mAP.
We speculate that either exchanging query information in CrossViT~\cite{DBLP:journals/corr/abs-2103-14899} or the shift and shuffle feature operations in Jigsaw~\cite{DBLP:journals/corr/abs-2102-04378} are ambiguous in such small $14\times14$ base feature maps, limiting the power of them for person search.
In contrast, our occluded attention is designed for small feature maps and obtains better performance, \ie, both $0.4\%$ gain in mAP and $1.0\%$ gain in top-1 score. Instead of sharing class tokens in different branches or shuffling channels of feature maps based on an individual instance, we effectively learn context information across different instances in a mini-batch, and differentiate the person from other people or the background to synthetically mimic occlusion.

{\noindent {\bf Comparison with feature augmentation.}}
Our method is related to previous augmentation strategies for person ReID, such as Batch DropBlock Network~\cite{DBLP:conf/iccv/DaiCGZT19}, Cutout~\cite{DBLP:journals/corr/abs-1708-04552} and Mixup~\cite{DBLP:conf/iclr/ZhangCDL18}. 
As presented in Table~\ref{tab:cross}, person search accuracy is not improved by using feature augmentation, simply augmenting feature patches with zeros.

\begin{table}[t]
\small
\begin{center}
\setlength{\tabcolsep}{7pt}
%\resizebox{\columnwidth}{!}
%{
\begin{tabular}{c|cc|cc}
\hline
\textbf{Method} & \textbf{Tokens} & \textbf{Feats} & \textbf{mAP} &\textbf{top-1}\\
\hline
Vanilla Attention & &  &52.9  &86.4 \\
CrossViT~\cite{DBLP:journals/corr/abs-2103-14899} &\Checkmark &  &49.9  &86.1 \\
Jigsaw~\cite{DBLP:journals/corr/abs-2102-04378}
&\Checkmark &  &51.9  &86.0\\
Batch DropBlock~\cite{DBLP:conf/iccv/DaiCGZT19}  & &\Checkmark  &52.7 &86.7 \\
Cutout~\cite{DBLP:journals/corr/abs-1708-04552}  & &\Checkmark  &53.2 &86.6 \\
Mixup~\cite{DBLP:conf/iclr/ZhangCDL18}  & &\Checkmark  &52.8 &86.6 \\
Occluded Attention &\Checkmark &  &\textbf{53.3}  &\textbf{87.4} \\
\hline
\end{tabular}
%}
% \vspace{-4mm}
\caption{Comparison of our attention mechanisms and other related modules. ``Tokens'' and ``Feats'' denote token-level enhanced attention and feature-level augmentation respectively. \label{tab:cross}
\vspace{-30pt}}
\end{center}
\end{table}

{\noindent {\bf Influence of occluded attention mechanism.}}
As discussed in Section~\ref{sec:occTransformer}, we use occluded attention to calculate discriminative person embeddings.  We evaluate the use of occluded attention (token mixup) and different scales in Table~\ref{tab:scale}.  Note, the top-1 score is improved from $86.4$ to $87.4$ with occluded attention and that multiple convolutional kernels for tokenization improve performance. Note that multiple convolutions do not increase the model size, since the feature maps $\mathcal{F}$ are channel-wise sliced for each scale. 

\begin{table}[t]
\small

\begin{center}
\setlength{\tabcolsep}{2pt}
%\resizebox{\columnwidth}{!}
%{
\begin{tabular}{c|cc|cc}
\hline
\textbf{Method} & \textbf{Token Mixup} & \textbf{Scales} & \textbf{mAP} &\textbf{top-1}\\
\hline
Vanilla Attention & &$\{1\times1\}$  &52.1  &85.3 \\
Vanilla Attention & &$\{3\times3\}$  &53.1  &86.0 \\
Vanilla Attention & &$\{1\times1,3\times3\}$  &52.9  &86.4 \\
Occluded Attention &\Checkmark &$\{1\times1\}$ &52.2  &86.5 \\
Occluded Attention &\Checkmark &$\{3\times3\}$ &52.5  &86.4 \\
Occluded Attention &\Checkmark &$\{1\times1,3\times3\}$  &\textbf{53.3}  &\textbf{87.4} \\
\hline
\end{tabular}
%}
\vspace{-10pt}
\caption{Comparison of our attention mechanisms and other related modules. ``Scales'' denotes the used convolutional kernels.\label{tab:scale}\vspace{-30pt}}
\end{center}
\end{table}

%------------------------------------------------------------------------
\begin{table}[t]
    \centering
    % \huge
    % \renewcommand\arraystretch{1.2}

    \resizebox{\columnwidth}{!}
    {
    \begin{tabular}{l|l|cc|cc}
    \hline
    \multicolumn{2}{c|}{\multirow{2}{*}{\textbf{Method}}} & \multicolumn{2}{c|}{\textbf{CUHK-SYSU}} & \multicolumn{2}{c}{\textbf{PRW}} \\ 
    \cline{3-6}
    \multicolumn{2}{c|}{} & \textbf{mAP}  & \textbf{top-1} & \textbf{mAP} & \textbf{top-1} \\ 
    \hline \hline
    \multirow{6}{*}{\rotatebox{90}{Two-step}} 
      & DPM~\cite{DBLP:conf/cvpr/ZhengZSCYT17} & - & - & 20.5 & 48.3 \\
      & MGTS~\cite{DBLP:conf/eccv/ChenZOYT18} & 83.0 & 83.7 & 32.6 & 72.1 \\
      & CLSA~\cite{DBLP:conf/eccv/LanZG18} & 87.2 & 88.5 & 38.7 & 65.0 \\
      & RDLR~\cite{DBLP:conf/iccv/HanYZTZGS19} & 93.0 & 94.2 & 42.9 & 70.2 \\
      & IGPN~\cite{DBLP:conf/cvpr/DongZST20} & 90.3 & 91.4 & \textbf{47.2} & 87.0 \\
      & TCTS~\cite{DBLP:conf/cvpr/WangMCSC20} & \textbf{93.9} & \textbf{95.1} & 46.8 & \textbf{87.5} \\ 
    \hline \hline
    \multirow{19}{*}{\rotatebox{90}{End-to-end}}
      & OIM~\cite{DBLP:conf/cvpr/XiaoLWLW17} & 75.5 & 78.7 & 21.3 & 49.9 \\
      & IAN~\cite{DBLP:journals/pr/XiaoXTHWF19} & 76.3 & 80.1 & 23.0 & 61.9 \\
      & NPSM~\cite{DBLP:conf/iccv/LiuFJKZQJY17} & 77.9 & 81.2 & 24.2 & 53.1 \\
      & RCAA~\cite{DBLP:conf/eccv/ChangHSLYH18} & 79.3 & 81.3 & - & - \\
      & CTXG~\cite{DBLP:conf/cvpr/YanZNZXY19} & 84.1 & 86.5 & 33.4 & 73.6 \\
      & QEEPS~\cite{DBLP:conf/cvpr/MunjalATG19} & 88.9 & 89.1 & 37.1 & 76.7 \\
      & HOIM~\cite{DBLP:conf/aaai/ChenZO0S20} & 89.7 & 90.8 & 39.8 & 80.4 \\
      & APNet~\cite{DBLP:conf/cvpr/ZhongWZ20} & 88.9 & 89.3 & 41.9 & 81.4 \\
      & BINet~\cite{DBLP:conf/cvpr/DongZST20a} & 90.0 & 90.7 & 45.3 & 81.7 \\
      & NAE~\cite{DBLP:conf/cvpr/ChenZYS20} & 91.5 & 92.4 & 43.3 & 80.9 \\
      & NAE+~\cite{DBLP:conf/cvpr/ChenZYS20} & 92.1 & 92.9 & 44.0 & 81.1 \\
      & DMRNet~\cite{DBLP:conf/aaai/HanZGSY21} & 93.2 & 94.2 & 46.9 & 83.3 \\
      & PGS~\cite{kim2021prototype} & 92.3 & \textbf{94.7} & 44.2 & 85.2 \\
      & AlignPS~\cite{DBLP:journals/corr/abs-2103-11617} & {93.1} & {93.4} & {45.9} & {81.9} \\
      & AlignPS+~\cite{DBLP:journals/corr/abs-2103-11617} & {94.0} & {94.5} & {46.1} & {82.1} \\
      & SeqNet~\cite{DBLP:conf/aaai/LiM21} & 93.8 & 94.6 & 46.7 & 83.4 \\
      & AGWF~\cite{han2021end} & 93.3 & 94.2 & \textbf{53.3} & \textbf{87.7} \\
      & COAT & \textbf{94.2} & \textbf{94.7} & \textbf{53.3} & 87.4 \\
    \cline{2-6}
      & AlignPS~\cite{DBLP:journals/corr/abs-2103-11617}+CBGM~\cite{DBLP:conf/aaai/LiM21} & {93.6} & {94.2} & {46.8} & {85.8} \\
      & AlignPS+~\cite{DBLP:journals/corr/abs-2103-11617}+CBGM~\cite{DBLP:conf/aaai/LiM21} & {94.2} & {94.3} & {46.9} & {85.7} \\
      & SeqNet+CBGM~\cite{DBLP:conf/aaai/LiM21} & \textbf{94.8} & \textbf{95.7} & 47.6 & 87.6 \\
      & COAT+CBGM & \textbf{94.8} &95.2  & \textbf{54.0} & \textbf{89.1}\\
    \cline{2-6}
      & HOIM$\dag$~\cite{DBLP:conf/aaai/ChenZO0S20} & - & - & 36.5 & 65.0 \\
      & NAE+$\dag$~\cite{DBLP:conf/cvpr/ChenZYS20} & - & - & 40.0 & 67.5 \\
      & SeqNet$\dag$~\cite{DBLP:conf/aaai/LiM21} & - & - & 43.6 & 68.5 \\
      & SeqNet+CBGM$\dag$~\cite{DBLP:conf/aaai/LiM21} & - & - & 44.3 & 70.6 \\
      & AGWF$\dag$~\cite{han2021end} & - & - & 48.0 & 73.2 \\
      & COAT$\dag$ &- &-  & {50.9} & {75.1}\\
      & COAT+CBGM$\dag$ &- &-  & \textbf{51.7} & \textbf{76.1}\\
    \hline 
    \end{tabular}
    }
    \caption{Comparison with the state-of-the-art methods. $\dag$ denotes the performance only evaluated on the multi-view gallery. Bold indicates highest score in the group. \label{tab:experiment}
    \vspace{-20pt}  
    }
\end{table}

\begin{figure*}[t]
    \centering
    \includegraphics[width=1.00\linewidth]{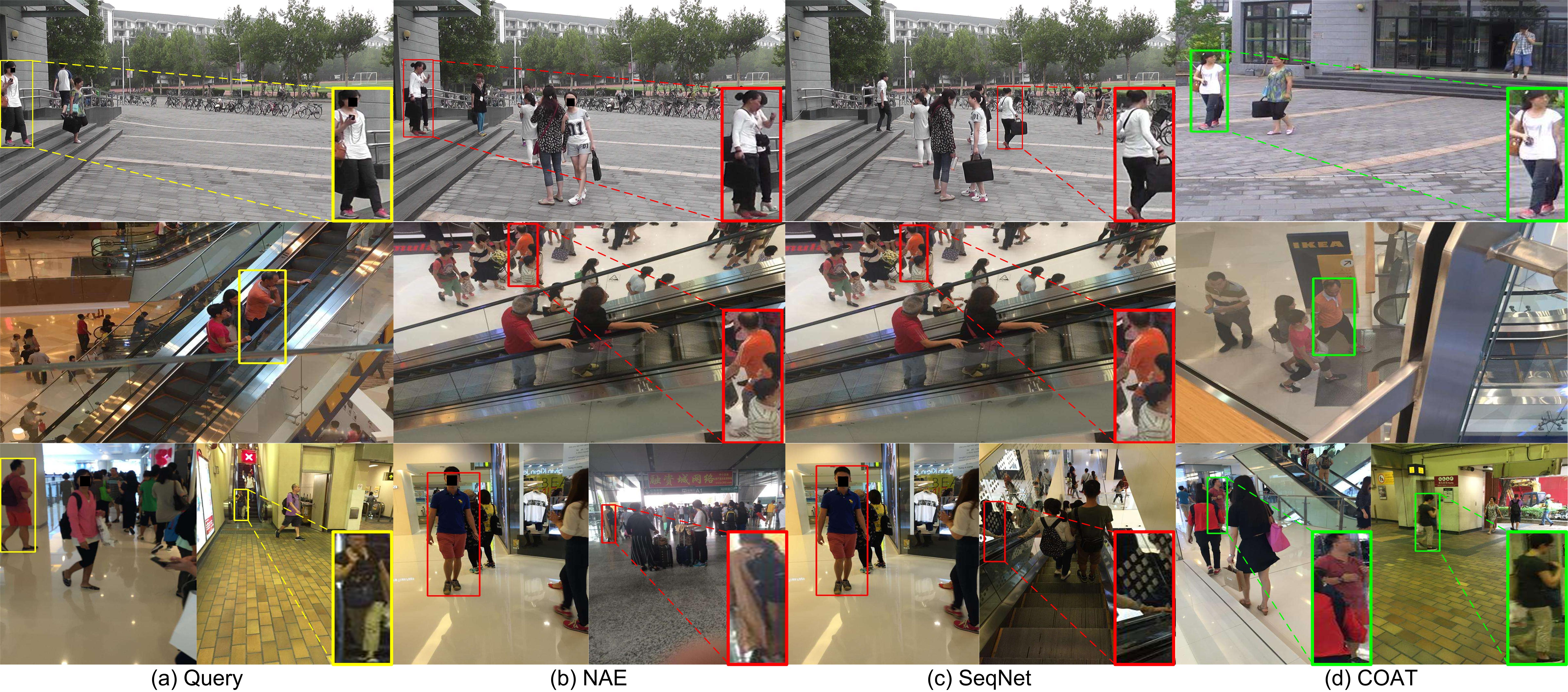}
    \vspace{-20pt}
    \caption{Qualitative examples of top-1 person search results of NAE~\cite{DBLP:conf/cvpr/ChenZYS20}, SeqNet~\cite{DBLP:conf/aaai/LiM21} and COAT on PRW (1st row) and CUHK-SYSU (2nd and 3rd rows) datasets, where small query, failure and correct cases are highlighted in yellow, red and green boxes respectively. \label{fig:visual}}
    \vspace{-4mm}
\end{figure*}

\subsection{Comparison with State-of-the-art}
\label{sec:experiment}
As presented in Table~\ref{tab:experiment}, we compare our COAT with state-of-the-art algorithms, including both two-step methods~\cite{DBLP:conf/cvpr/ZhengZSCYT17,DBLP:conf/eccv/ChenZOYT18,DBLP:conf/eccv/LanZG18,DBLP:conf/iccv/HanYZTZGS19,DBLP:conf/cvpr/DongZST20,DBLP:conf/cvpr/WangMCSC20} and end-to-end methods~\cite{DBLP:conf/cvpr/XiaoLWLW17,DBLP:journals/pr/XiaoXTHWF19,DBLP:conf/iccv/LiuFJKZQJY17,DBLP:conf/eccv/ChangHSLYH18,DBLP:conf/cvpr/YanZNZXY19,DBLP:conf/cvpr/MunjalATG19,DBLP:conf/aaai/ChenZO0S20,DBLP:conf/cvpr/ZhongWZ20,DBLP:conf/cvpr/DongZST20a,DBLP:conf/cvpr/ChenZYS20,kim2021prototype,DBLP:journals/corr/abs-2103-11617,DBLP:conf/aaai/LiM21,han2021end}, on two datasets.

\begin{figure}[t]
\setlength{\abovecaptionskip}{1mm}
    \centering
    \begin{subfigure}[b]{0.49\linewidth}
           \centering
           \includegraphics[width=\linewidth]{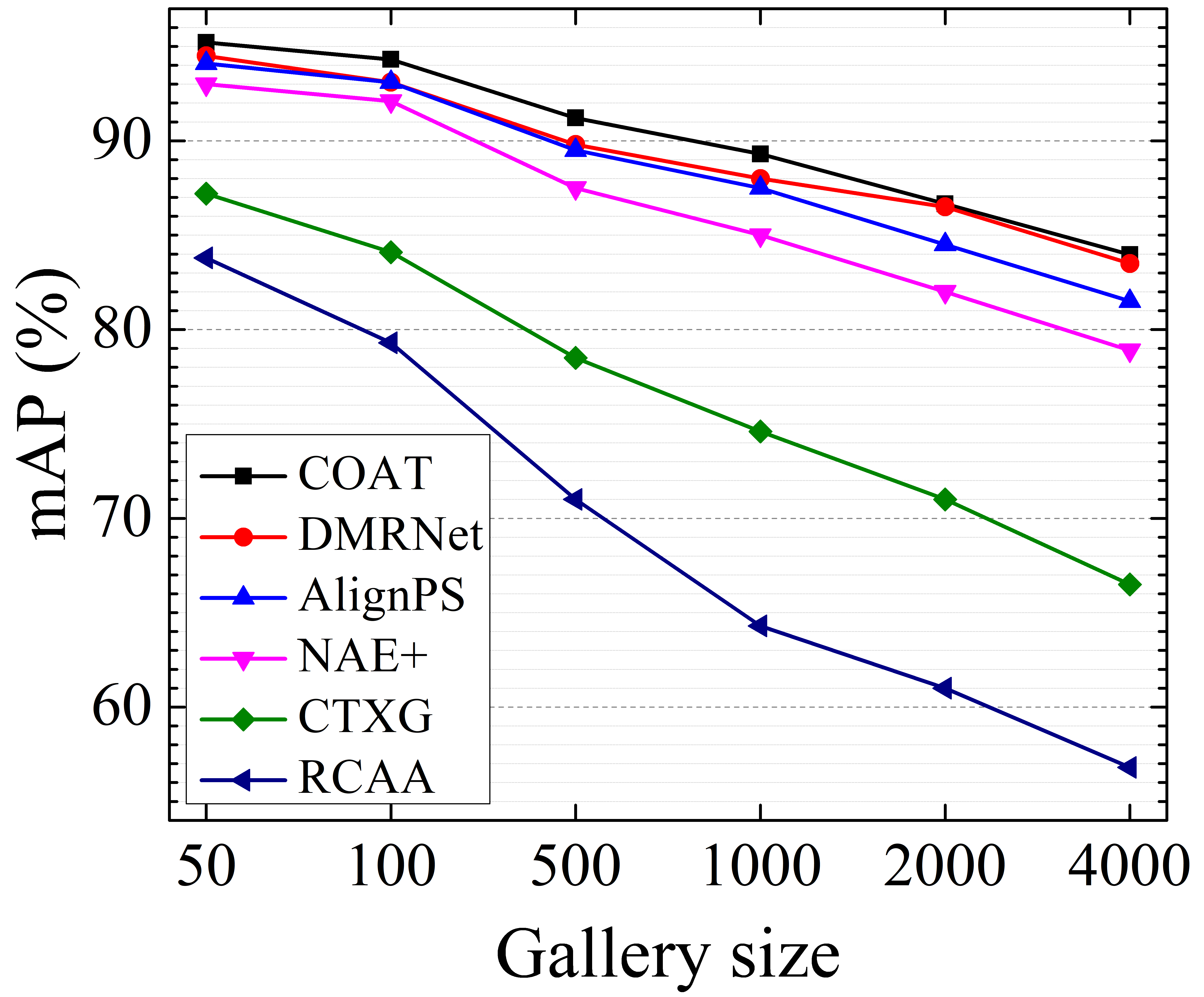}
            \caption{End-to-end models}
            \label{subfig:cuhk-one-step}
    \end{subfigure}
    \begin{subfigure}[b]{0.49\linewidth}
            \centering
            \includegraphics[width=\linewidth]{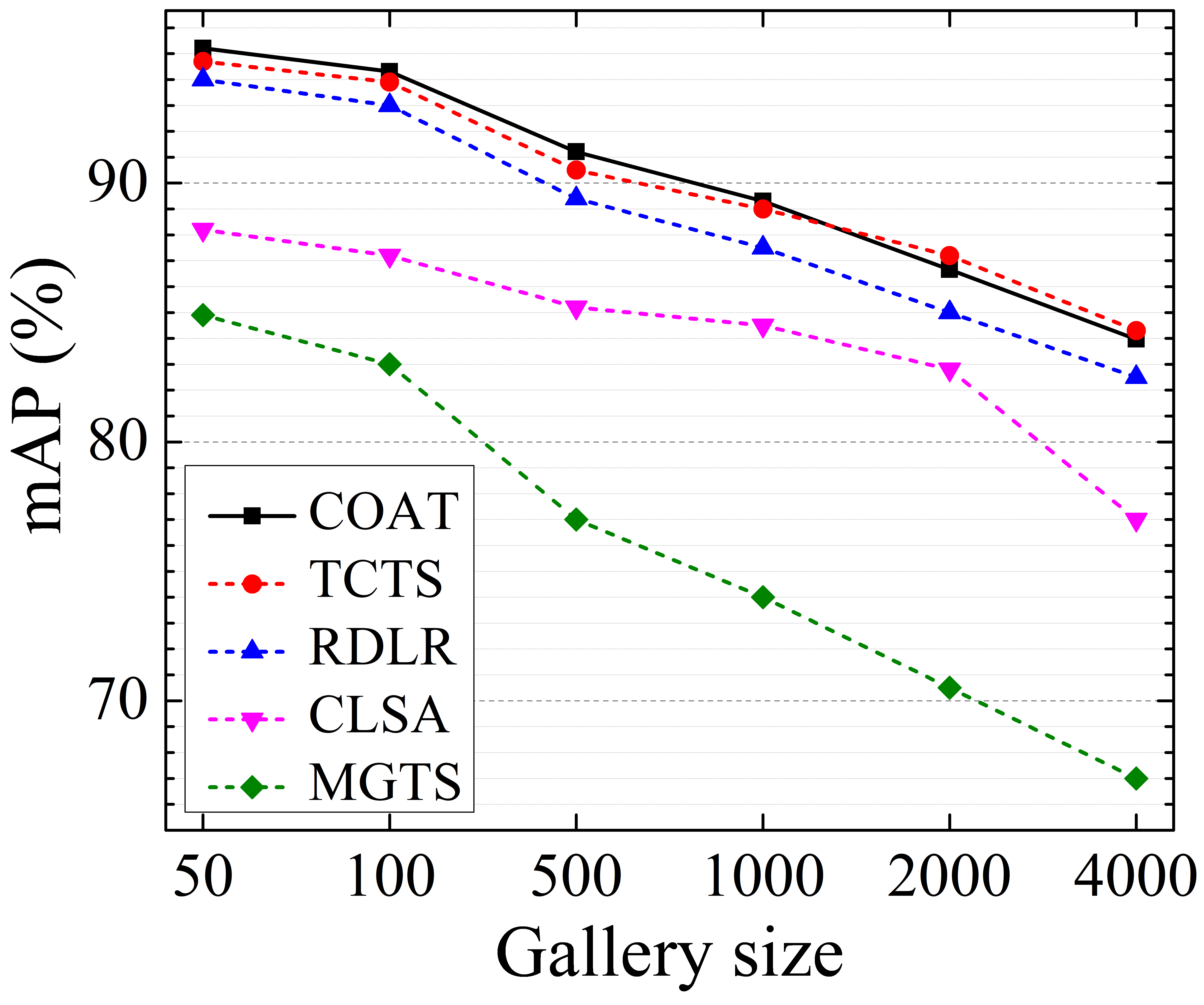}
            \caption{Two-step models}
            \label{subfig:cuhk-two-step}
    \end{subfigure}
    \caption{Comparison with (a) end-to-end models and (b) two-step models on CUHK-SYSU with different gallery sizes.\label{fig:cuhk-gallery-size}}
    \vspace{-20pt}
\end{figure}

{\noindent {\bf Results on CUHK-SYSU.}} 
With the gallery size of $100$, our method achieves the best $94.2\%$ mAP and comparable $94.7\%$ \mbox{top-1} scores compared to the best two-step method TCTS~\cite{DBLP:conf/cvpr/WangMCSC20} with explicitly trained bounding box and ReID feature refinement modules. 
Among end-to-end methods, our method performs better than state-of-the-art AlignPS+~\cite{DBLP:journals/corr/abs-2103-11617} with a multi-scale anchor-free representation~\cite{DBLP:conf/iccv/TianSCH19}, SeqNet~\cite{DBLP:conf/aaai/LiM21} with two-stage refinement and AGWF~\cite{han2021end} with part classification based sub-networks. The results indicate the effectiveness of our cascaded multi-scale representation. Using the post-processing operation Context Bipartite Graph
Matching (CBGM)~\cite{DBLP:conf/aaai/LiM21}, both mAP and top-1 scores of our method can be further improved slightly. 
For a comprehensive evaluation, as shown in Figure~\ref{fig:cuhk-gallery-size}, we compare mAP scores of competitive methods as we increase gallery size. Since it is challenging to consider more distracting people in the gallery set, the performance of all compared methods is reduced as the gallery size increases. However, our method consistently outperforms all the end-to-end methods and the majority of two-step methods. When the gallery size is larger than $1,000$, our method performs slightly worse than the two-step TCTS~\cite{DBLP:conf/cvpr/WangMCSC20}. 

{\noindent {\bf Results on PRW.}} 
Although the PRW dataset~\cite{DBLP:conf/cvpr/ZhengZSCYT17} is more challenging, with less training data but larger gallery size, than the CUHK-SYSU dataset~\cite{DBLP:conf/cvpr/XiaoLWLW17}, the results show a similar trend. Our method achieves comparable performance as AGWF~\cite{han2021end} and a significant gain of $6.7\%$ mAP and $4.0\%$ top-1 scores than SeqNet~\cite{DBLP:conf/aaai/LiM21}. 
DMRNet~\cite{DBLP:conf/aaai/HanZGSY21} and AlignPS~\cite{DBLP:journals/corr/abs-2103-11617} leverage stronger object detectors, such as RetinaNet~\cite{DBLP:conf/iccv/LinGGHD17} and FCOS~\cite{DBLP:conf/iccv/TianSCH19}, than the Faster R-CNN~\cite{DBLP:journals/pami/RenHG017} in our method, but still achieve inferior performance. 
Further, we compare performance on PRW's multi-view gallery (see the group marked by $\dag$ in Table~\ref{tab:experiment}). Our method outperforms existing methods in terms of both mAP and Top-1 scores with a clear margin.
We attribute this to our cascaded transformer structure which generates more discriminative ReID features, especially in the cross-camera setting with significant pose/viewpoint changes.

{\noindent {\bf Qualitative results.}} 
Some example person search results on two datasets are shown in Figure~\ref{fig:visual}. Our method can deal with cases of slight/moderate occlusion and scale/pose variations, while other state-of-the-art methods such as SeqNet~\cite{DBLP:conf/aaai/LiM21} and NAE~\cite{DBLP:conf/cvpr/ChenZYS20} fail in these scenarios.

{\noindent {\bf Efficiency comparison.}} 
We compare our efficiency with three representative end-to-end networks including NAE~\cite{DBLP:conf/cvpr/ChenZYS20}, AlignPS~\cite{DBLP:journals/corr/abs-2103-11617} and SeqNet~\cite{DBLP:conf/aaai/LiM21} which have publicly released source code. We evaluate the methods with the same scale test images and on the same GPU.

From Table~\ref{tab:speed}, we compare the number of parameters, the multiply–accumulate operations (MACs), and the running speed in frames per second (FPS). Our method has lower computational complexity and slightly slower speed than other compared methods, but achieved $+6.6\%$ and $+4.0\%$ gains in mAP and top-1 accuracy respectively. In contrast to~\cite{DBLP:conf/iclr/DosovitskiyB0WZ21,DBLP:journals/corr/abs-2102-04378}, we employ only one encoder layer in our transformers and use multi-scale convolutions to reduce the number of channels before tokenization, increasing COAT's efficiency.

\begin{table}[t]
% \small
\begin{center}
\setlength{\tabcolsep}{4pt}
    \resizebox{\columnwidth}{!}
    {
\begin{tabular}{c|ccc|cc}
\hline
\textbf{Method} & \textbf{Params(M)} & \textbf{MACs(G)} & \textbf{FPS}  &\textbf{mAP} &\textbf{top-1}\\
\hline
NAE~\cite{DBLP:conf/cvpr/ChenZYS20} & 33.43 &287.35 &14.48 &43.3  &80.9 \\
AlignPS~\cite{DBLP:journals/corr/abs-2103-11617} & 42.18 &189.98  &16.39 & 45.9 & 81.9\\
SeqNet~\cite{DBLP:conf/aaai/LiM21} & 48.41 & 275.11 &12.23 &46.7  &83.4 \\
\hline
COAT & 37.00 & 236.29 &11.14 &\textbf{53.3}  &\textbf{87.4} \\
\hline
\end{tabular}
}
\vspace{-20pt} % this spacing needs to be balanced with the one after the caption.  Ask Chris to do this.
\caption{Comparison of person search efficiency.
\label{tab:speed}
\vspace{-20pt}}
\end{center}
\end{table}

\section{Conclusion}
We have developed a new Cascade Occluded Attention Transformer (COAT) for end-to-end person search. Notably, COAT learns a discriminative coarse-to-fine representation for both person detection and person ReID via a cascade transformer framework. Meanwhile, the occluded attention mechanism synthetically mimics occlusions from either foreground or background objects. COAT outperforms state-of-the-art methods, which we hope will inspire more research into transformer-based person search methods. 

{\noindent {\bf Ethical considerations.}} 
Like most technologies, person search methods may have societal benefits and negative impacts. How the technology is employed is critical. For example, person search can identify persons of interest to aid law enforcement and counter-terrorism operations. However, the technology should only be used in locations where an expectation of privacy is waived by entering those locations, such as public areas, airports, and private buildings with clear signage. These systems should not be employed without probable cause, or by unjust governments that seek to acquire ubiquitous knowledge of the movements of all of their citizens to enable persecution and repression.  

For comparability, this research uses human subjects imagery collected in prior works. CUHK-SYSU~\cite{DBLP:conf/cvpr/XiaoLWLW17} was collected from ``street snaps'' and ``movie snapshots'', while PRW~\cite{DBLP:conf/cvpr/ZhengZSCYT17} was collected with video cameras in a public area of a university campus.  No mention is made in either paper of review by an ethical board (\eg, an Institutional Review Board), but these papers were published before this new standard was established at CVPR or most major AI conferences. Our preference would be to work with ethically collected person search datasets, and we would welcome a public disclosure from the authors of their ethical compliance. We believe the community should focus resources on developing ethical person search datasets and phase out the use of legacy, unethically collected datasets.

{\noindent {\bf Acknowledgement.}} 
This material is based upon work supported by the United States Air Force under Contract No. FA8650-19-C-6036. Any opinions, findings and conclusions or recommendations expressed in this material are those of the author(s) and do not necessarily reflect the views of the United States Air Force.

%%%%%%%%% REFERENCES
{\small
\bibliographystyle{ieee_fullname}
\bibliography{egbib}
}

\end{document}